# Automatic Detection, Positioning and Counting of Grape Bunches Using Robots

Xumin Gao
*The school of computer science*
*University of Lincoln*
Lincoln, UK
25766099@students.lincoln.ac.uk

*Abstract*—In order to promote agricultural automatic picking and yield estimation technology, this project designs a set of automatic detection, positioning and counting algorithms for grape bunches, and applies it to agricultural robots. The Yolov3 detection network is used to realize the accurate detection of grape bunches, and the local tracking algorithm is added to eliminate relocation. Then it obtains the accurate 3D spatial position of the central points of grape bunches using the depth distance and the spatial restriction method. Finally, the counting of grape bunches is completed. It is verified using the agricultural robot in the simulated vineyard environment. The project code is released at: https://github.com/XuminGaoGithub/Grape_bunches_count_using_robots.

## I. Introduction

At present, the development of agricultural automation is still in a slow state. Most agricultural work still mainly depends on human labor, and its work efficiency is low. In addition, with the continuous development of urbanization, the shortage of labor resources in the field of agriculture are becoming more and more serious. In recent years, with the development of robot and sensor technology, the application of robots in agriculture has become a feasible method. Using agricultural robots to automatically pick fruits and estimate yield is one of the hot research topics. Z. De-An et al. [1] designed an apple harvesting robot with the 5 DOF PRRRP structure's manipulator. Y. Xiong et al. [2] designed a strawberry picking robot, which achieved good results in the scenario where strawberries were partially surrounded or totally isolated. Accurate detection, positioning and counting algorithms are the key technologies for agricultural robots to realize automatic yield estimation and picking of fruits.

Therefore, this project combines Yolov3 [3] detection network, local tracking algorithm, depth information from Kinect sensor and spatial restriction method to automatically detect, locate and count grape bunches based on the simulated Thorvald robot and vineyard environment developed by LCAS. At the same time, the robot can automatically navigate around the vineyard through topology map and movebase package in ROS. It can be extended to any kind of agricultural robots to automatically detect, locate and count different kinds of fruits.

## II. Related work

The existing work about detection, positioning and counting of grape bunches are mainly divided into two aspects. The first aspect of work is based on 2D images and the second aspect of work is based on 3D point clouds.

### A. The work based on 2D images

The work based on 2D images are mainly divided into two methods. The first method directly uses color threshold to extract grape bunches, or extracts image features (color, geometry and texture) combined with traditional machine learning classifier to detect grape bunches. M.J. Reis et al. [4] recognized the red grapes and white grapes by distinguishing the RGB value in the image. S. Liu et al. [5] extracted the color, texture and morphological features of grape bunches in the image, and then trained the model that can detect grape bunches using SVM [6] classifier. This method relies heavily on the selected image features, so the detection effect is poor, and it is difficult to distinguish multiple targets with partially overlapping areas in the image. The second method uses deep learning based on convolutional neural network to detect or segment grape bunches. A.S. Aguiar et al. [7] collected grape bunches dataset at different growth stages, and trained and tested it using SSD-MobileNet_V1 and SSD-Inception_V2 respectively. L. Ghiani et al. [8] used Mask R-CNN [9] to train the collected grape bunches dataset and achieved good test results. Compared with the traditional image feature extraction and classifier learning, the detection effect based on deep learning method is more robust and more adaptive to different scenarios.

In general, the work based on 2D images has great limitations. It can only detect and count the number of targets from a single image, and can not estimate the 3D spatial coordinates of the grape bunches.

### B. The work based on 3D point clouds

Most of the work based on 3D point clouds focuses on the 3D reconstruction of grape berries to estimate the corresponding volume and mass. In recent years, some new 3D pose estimation and counting of grape bunches have been derived on the basis of 3D point clouds. W. Yin et al. [10] used Mask R-CNN to detect and segment the grape bunches based on the binocular stereo camera. Then extracting point clouds of the grape bunches according to the segmented image, and using RANSAC algorithm [11] to post-process point clouds. Finally the pose estimation of the grape bunches is realized. A.K. Nellithimaru et al. [12] used SLAM technology combined with 3D reconstruction, image instance segmentation to count grape berries.

However, the biggest limitation of the work based on 3D point clouds is that it will occupy huge computing resources and can not realize real-time processing. In addition, the computing capability of robot's platform is not high normally.

## III. SYSTEM ARCHITECTURE

This project designed a set of automatic detection, positioning and counting algorithms for grape bunches, and applied it to agricultural robots. Firstly, the robot uses Yolov3 to detect grape bunches, and uses a simple local tracking algorithm to track the central points of the grape bunches. Then obtaining the depth distance of the central points of the grape bunches by registering Kinect's color image and depth image, and calculating target's 3D position in world coordinate frame. Meanwhile, using spatial restriction method to filter out positioning noises. Finally, the robot can complete the counting task and publish the 3D poses of all grape bunches to the Rviz interface. For the navigation, the robot realizes autonomous navigation around the grape trellis using the pre-established topology map and movebase package in ROS.

### A. Detection

In order to train the Yolov3 model that can detect grape bunches, the robot randomly collected 200 images from the simulated vineyard, then labelling them manually. To prevent over fitting of the model, the dataset is augmented by using randomly adding noise, changing brightness, rotation, clipping and translation to original images. The final size of dataset is 1000 images.

The dataset is divided into training dataset and test dataset according to 8:2. Using Yolov3 network to train detection model of grape bunches. The maximum number of iterations is set to 20000. The training is completed in the local laptop configured with NVIDIA GTX1060 GPU. The average loss change curve during the training process is shown in Fig. 1.

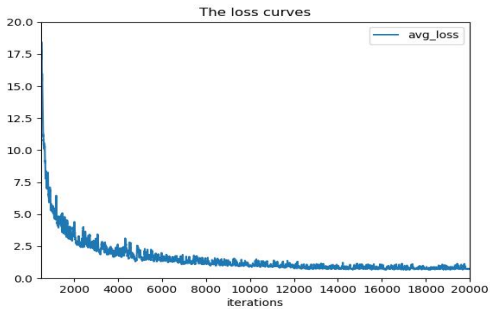

**Fig. 1** The training average loss curve

It can be seen from Fig. 1 that after 14000 iterations, the average loss tends to be stable, so the weight with 14000 iterations is used as the final grape bunches detection model. In this project, when the grape bunches are detected by Yolov3, the corresponding bounding boxes are published through ROS topic.

### B. Tracking

In order to eliminate the re-identification of grape bunches and then lead to relocation when robot is moving, calculating euclidean distance to track central points of grape bunches from different camera frames. The principle diagram of tracking of central points of grape bunches between consecutive frames is shown in Fig. 2.

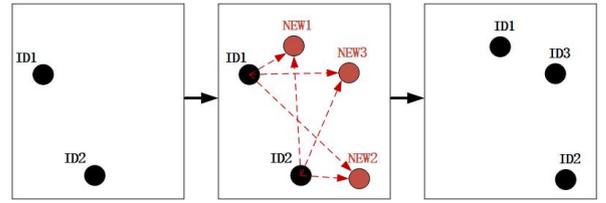

**Fig. 2** The tracking principle between continuous frames

In Fig. 2, there are two black points numbered ID1 and ID2 in the first frame image, and three red points numbered NEW1, NEW2, and NEW3 in the second frame image. Calculating the euclidean distance between each original point and the new point. Then associating and matching the points of two consecutive frames through searching the minimum distance. Unmatched points will be assigned a new ID. Thus, it can be seen that there are two points (ID1 and ID2) matching from the first frame to the second frame image, and a new ID (ID3) is assigned. In addition, it needs to set the maximal disappeared number of frames $N$ when an old ID disappears in the image. If it cannot match with the existing ID within $N$ frames, it will be deregistered.

### C. Positioning

The 3D spatial positioning of grape bunches is divided into two steps. The first step is to register the color image and depth image from Kinect, so as to obtain the depth distance value of the central points of the grape bunches from the registered depth image. Considering the *depth_image_proc* node in ROS needs to process a large amount of dense data in the process of registration, which will lead to performance loss in terms of timeliness. This project adopts a slim depth image registration library [13], which can quickly complete the registration of color image and depth image. The second step is to transform the central points of the grape bunches from the image coordinate frame to the world coordinate frame using the principle of pinhole camera model and TF tree in ROS, so that it can obtain the corresponding 3D spatial position of grape bunches.

### D. Spatial restriction method

When the robot moves, the 3D spatial coordinates of the central point of the same grape bunch obtained from 2D image from different perspectives are not exactly equal, which will lead to relocation of the grape bunches. Assuming that the grape bunch is a cylindrical model in 3D space, as shown in Fig. 3, the 3D spatial position of the central point of the grape bunches obtained from 2D image from different perspectives must be within the cylinder.

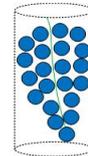

**Fig. 3** The Spatial cylindrical model of grape bunch

To this end, adding the space restriction method for positioning grape bunches. At first, saving the 3D space coordinates $(x1, y1, z1)$ of the central point of the first detected grape bunch. When the second grape bunch is detected that the 3D space coordinates of central point is $(x2, y2, z2)$, it makes the difference between $(x1, y1, z1)$ and $(x2, y2, z2)$, as shown in Equation (1).

$$diff = \frac{|x_2 - x_1| + |y_2 - y_1| + |z_2 - z_1|}{3} \quad (1)$$

Then, comparing *diff* with the *threshold* which is set according to the length, width and height of the cylindrical model of the grape bunch. If $diff > threshold$, the second detected grape bunch is regarded as a new target, and add it to the counting list. On the contrary, if $diff \leq threshold$, the second detected grape bunch is considered to be relocated with the first detected grape bunch, it will not be added to the counting list. By analogy, whenever the robot detects a new grape bunch, it will compare the 3D spatial coordinates of the central point with that of all existing grape bunches, so as to determine whether the new detected grape bunch is added to the counting list.

*E. Navigation*

In this project, the robot can automatically navigate around the grape trellis using the topology map and combining with movebase package in ROS. The best perceptual range for Kinect is 0.5 ~ 4.5m, therefore the robot only performs the counting task when moving close to the grape trellis, which can eliminate some noises and get more accurate counting results.

*F. Counting*

On the basis of A-E, the robot can realize the automatic detection, positioning and counting of all grape bunches in the simulated vineyard environment.

## IV. EVALUATION

*A. Detection*

The trained Yolov3 model is used to evaluate the detection accuracy and speed of grape bunches on the test dataset. The results of mAP (mean Average Precision) and mFPS (mean Frames Per Second) are shown in Table I.

**Table I.** The mAP and mFPS of trained Yolov3 model

| mAP | mFPS |
| --- | --- |
| 0.9567 | 22 |

At the same time, making comparison between the traditional HSV color extraction detection and Yolov3 detection. The comparison results are shown in Fig. 4.

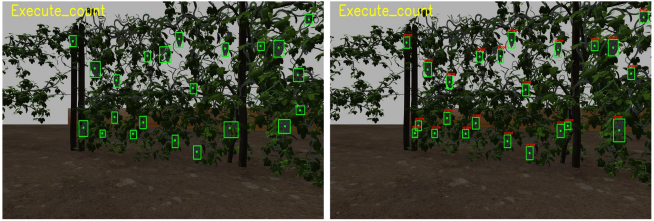

a）HSV color extraction detection     b）Yolov3 detection

**Fig. 4** The comparison results of HSV color extraction detection and Yolov3 detection

As can be seen from Fig. 4, compared with the traditional HSV color extraction detection method, Yolov3 detection can effectively distinguish and detect grape bunches with partial overlapping areas or extremely close to each other in the image.

*B. Tracking*

In order to verify the effect of the tracking algorithm, collecting the detection and tracking images when the robot moves from time $t$ to $t + n$. The results are shown in Fig. 5.

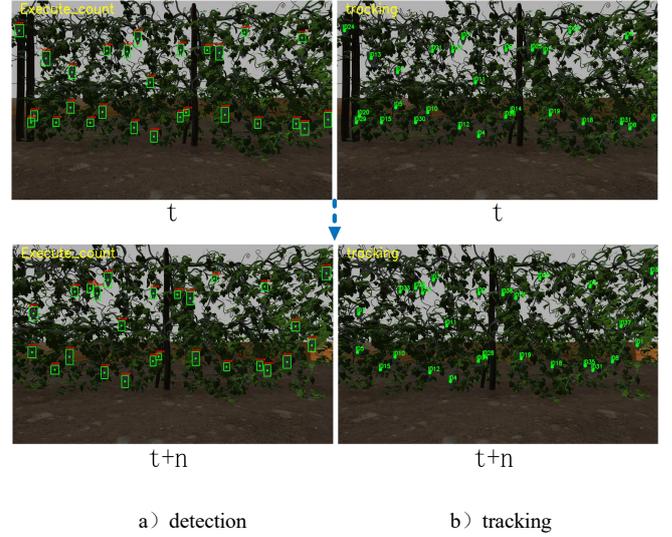

a）detection     b）tracking

**Fig. 5** The detection and tracking of grape bunches

As can be seen from Fig. 5, it can effectively track the isolated grape bunches. For the close grape bunches, it is easy to cause ID matching disorder, but this error can be eliminated through the subsequent 3D positioning. Also, please note that the ID number here is not the counting result.

*C. Positioning*

  *a) Registration*

When the robot performs the counting task, collecting the color image and the registered depth image with the detection information of grape bunches at a certain time $t$. The results are shown in Fig. 6.

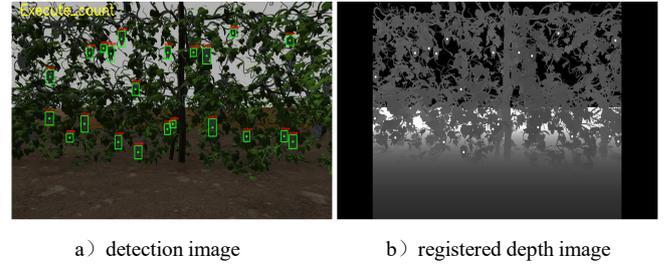

a）detection image     b）registered depth image

**Fig. 6** The detection image and registered depth image

In Fig. 6, the detection image contains the detection bounding boxes and central points (green solid point) of grape bunches with effective depth distance. The registered depth image contains the central points (white solid point) of the corresponding grape bunches. It can be seen that the central points of the detected grape bunches basically coincide between the color image and the registered depth image. Therefore, the registration algorithm has achieved good result.

  *b) 3D spatial position*

In this project, the 3D poses of grape bunches are published to the Rviz interface for visualization. Each of 3D spatial positions of grape bunches will be visualized as a sphere marker in the Rviz interface.

## D. Counting

When the robot performs the counting task, capturing the counting states of grape bunches at the beginning, middle and end times respectively, as shown in Fig. 7.

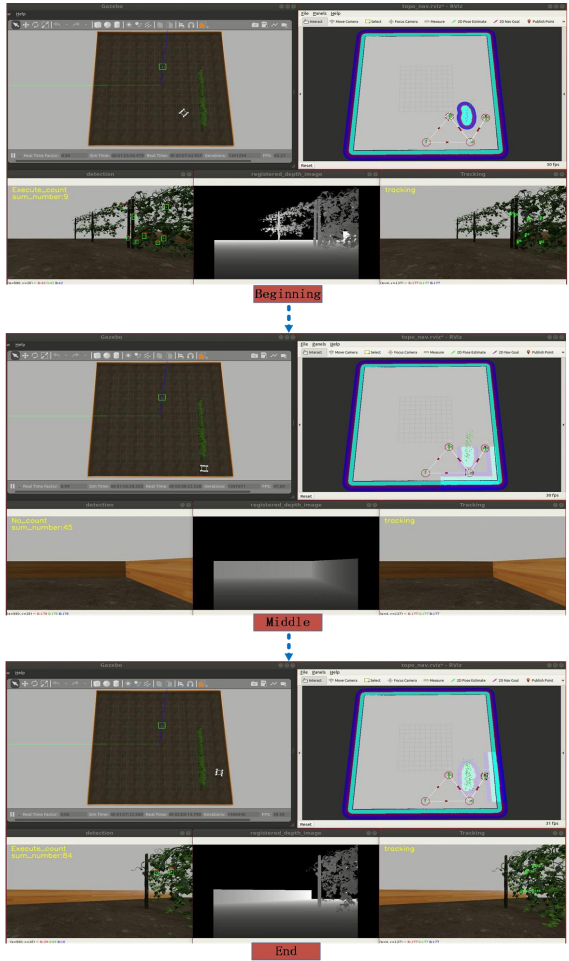

**Fig. 7** The counting states of grape bunches at different times

As can be seen from Fig. 7, the robot can accurately complete the automatic detection, positioning and counting of grape bunches in the simulated vineyard environment.

It is worth noting that the counting number of grape bunches in this task is 84, but actually this number is not fixed. Although the robot automatically navigates according to the same topology node every time when it performs counting task, the local path planning is not exactly the same every time. This also leads to the fact that the images obtained by the robot's Kinect are not exactly the same every time. Therefore, the results of counting are slightly different every time. Using the robot to perform counting task for ten times, and the results are shown in the table II.

**Table II**. The counting results of ten times

| Number | The result of counting |
|---|---|
| 1 | 84 |
| 2 | 90 |
| 3 | 86 |
| 4 | 89 |
| 5 | 77 |
| 6 | 78 |
| 7 | 79 |
| 8 | 89 |
| 9 | 88 |
| 10 | 87 |
| **The average counting value** | **85** |

Calculating the average of them, the average counting value of the grape bunches is 85.

## V. CONCLUSION

This project designs a set of accurate automatic detection, positioning and counting algorithms for grape bunches. The validation experiment is carried out in the simulated vineyard environment. Yolov3 has achieved good detection performance, it can effectively distinguish and detect grape bunches with partial overlapping areas or extremely close to each other in the image. The local tracking algorithm has poor tracking effect for the close grape bunches. In the future, it can be considered using Deepsort [14] to get the more robust tracking. The positioning of grape bunches has also achieved good results, but only the 3D spatial coordinates of the central of the grape bunches are calculated. In the future, it can be considered adding the segmentation algorithm to replace detection, so as to obtain a set of 3D spatial coordinates of the grape bunches instead of one central point.

ACKNOWLEDGEMENTS

This work was supported by the Engineering and Physical Sciences Research Council [EP/S023917/1], the AgriFoRwArdS CDT, and the University of Lincoln.